\makeatletter\def\graphicscache@inhibit{true}\makeatother

\documentclass[letterpaper, 10 pt, conference]{ieeeconf}  %

\IEEEoverridecommandlockouts                              %

\usepackage[utf8]{inputenc}

\usepackage[hidelinks]{hyperref}
\usepackage[style=ieee,hyperref,natbib=true,backend=bibtex,firstinits,doi=false,%
     mincitenames=1,maxcitenames=2,maxbibnames=99,sorting=none,terseinits=false,hyperref=true]{biblatex}
\bibliography{references.bib}
\renewbibmacro*{bbx:savehash}{}%
\defbibheading{bibliography}[\bibname]{\section*{References}}

\usepackage{url}
\usepackage{graphicx}
\usepackage[usenames,dvipsnames]{xcolor}
\usepackage[draft,nomargin,inline]{fixme}
\fxsetup{theme=color}

\usepackage{amsmath}
\usepackage{amssymb}
\usepackage{textcomp}
\usepackage{siunitx}

\usepackage{booktabs}
\usepackage{threeparttable}
\usepackage{multirow}

\usepackage[capitalize]{cleveref}

\usepackage{tikz}
\usetikzlibrary{arrows}
\usetikzlibrary{positioning,calc}
\usetikzlibrary{decorations.pathreplacing}
\usetikzlibrary{decorations.markings}
\usetikzlibrary{fit}
\usetikzlibrary{shapes.callouts}
\usetikzlibrary{shapes.geometric}
\usetikzlibrary{matrix}

\usepackage{pgfplotstable}
\usepackage{pgfplots}
\pgfplotsset{compat=1.15}
\usepgfplotslibrary{groupplots}
\usepgfplotslibrary{units}
\usepgfplotslibrary{statistics}

\usepackage{blindtext}

\usepackage[export]{adjustbox}

\usepackage{placeins}
\usepackage{dblfloatfix}

\usepackage{blindtext,letltxmacro,xcolor,xparse}
\LetLtxMacro{\blindtextblindtext}{\blindtext}
\LetLtxMacro{\blindtextBlindtext}{\Blindtext}

\RenewDocumentCommand{\blindtext}{O{\value{blindtext}}}{%
  \begingroup\color{gray}\blindtextblindtext[#1]\endgroup
}

\IfFileExists{graphicscache.sty}{\usepackage{graphicscache}}

\makeatletter
\newcommand\coord{\the\tikz@lastxsaved,\the\tikz@lastysaved}
\makeatother

\title{\LARGE \bf
Low-Latency Immersive 6D Televisualization with Spherical Rendering
}

\author{Max Schwarz$^{*}$ and Sven Behnke%
\thanks{$^{*}$Both authors are with the Autonomous Intelligent Systems group of University of Bonn, Germany; {\tt schwarz@ais.uni-bonn.de}}%
}

\begin{document}

\maketitle

\begin{abstract}

We present a method for real-time stereo scene capture and remote VR
visualization that allows a human operator to freely move their head
and thus intuitively control their perspective during teleoperation.
The stereo camera is mounted on a 6D robotic arm, which follows the operator's head pose.
Existing VR teleoperation systems either induce high latencies on head
movements, leading to motion sickness, or use scene reconstruction methods
to allow re-rendering of the scene from different perspectives, which
cannot handle dynamic scenes effectively.
Instead, we present a decoupled approach which renders captured camera images
as spheres, assuming constant distance. This allows very fast re-rendering on
head pose changes while keeping the resulting temporary distortions during head translations small.
We present qualitative examples, quantitative results in the form of lab
experiments and a small user study, showing that our method outperforms
other visualization methods.

\end{abstract}

\section{Introduction}

There are many ways to control robotic systems, from remote control over
teleoperation to full autonomy. Teleoperation is highly relevant and valuable
to the robotic research community, first of all because it allows to address
tasks that are impossible to solve using state-of-the-art autonomous control
methods---the human intellect and creativity in solving problems
and reacting to unforeseen events is unmatched. Furthermore, teleoperation
can generate training examples for improving autonomy in the future.
There are also immediate benefits for the economy and daily life: Teleoperation
can allow humans to work remotely in potentially dangerous environments, which is
highly desirable in situations like the current COVID-19 pandemic.
This interest is embodied in new robotics competitions like the ANA Avatar
XPRIZE Challenge\footnote{https://www.xprize.org/prizes/avatar} and a multitude
of other past and ongoing competitions~\citep{krotkov2017darpa,kitano1999robocup}.

One important, if not the most important feature of a teleoperation system is
its capability to display the remote environment to the human operator in a
convincing, immersive way. Maintaining full situational awareness is necessary
to both induce security and confidence for the operator as well as to solve
tasks efficiently and robustly.

We present a real-time scene capture and visualization system for teleoperated
robots, which gives the human operator wearing a head-mounted display
full 6D control of the first-person camera perspective. In summary, our contributions include:

\begin{enumerate}
 \item A robotic head assembly equipped with a 180° 4K stereo camera pair, capable of 6D
    movement with low latency;
 \item a combination of calibration steps for camera intrinsics and extrinsics, hand-eye, and VR calibration; and
 \item an efficient rendering method that decouples the VR view from the
    latency-afflicted robot head pose, providing a smooth low-latency
    VR experience at essentially unlimited frame rate.
\end{enumerate}

\begin{figure}
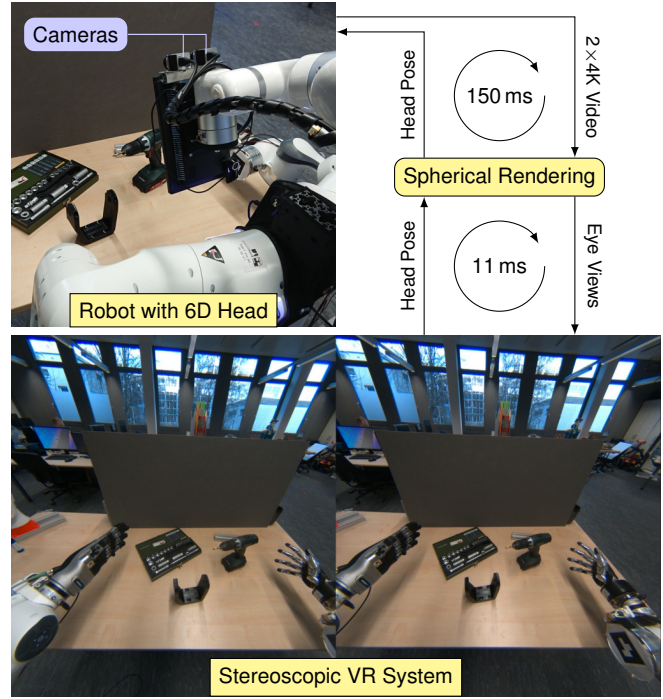

 \centering

 \begin{tikzpicture}[
   font=\sffamily\footnotesize,
   mod/.style={draw=black,fill=yellow!50,rounded corners,align=center},
   l/.style={font=\sffamily\scriptsize},
  ]
  \node[anchor=south west,inner sep=0] (head) {\includegraphics[width=.5\linewidth]{images/teaser/teaser.jpg}};
  \begin{scope}[x={($ (head.south east) - (head.south west) $ )},y={( $ (head.north west) - (head.south west)$ )}, shift={(head.south west)}]
   \node[fill=blue!20,rounded corners] (camlabel) at (0.2,0.9) {Cameras};
   \draw[draw=blue!20] (camlabel) -| (0.60,0.84);
   \draw[draw=blue!20] (camlabel) -| (0.53,0.83);
  \end{scope}

  \node[anchor=north west,inner sep=0] (vr) at (0,-0.1) {\includegraphics[width=\linewidth]{images/vr_view.jpg}};

  \node[mod] (rendering) at (6.5,2) {Spherical Rendering};

  \draw[-latex] ($(rendering.south east)+(-0.4,0)$) -- node[l,pos=0.5,rotate=-90,above] {Eye Views} (\coord|-vr.north);
  \draw[latex-] ($(rendering.south west)+(+0.4,0)$) -- node[l,pos=0.5,rotate=90,above] {Head Pose} (\coord|-vr.north);

  \draw[-latex] ($(head.north east)+(0,-0.2)$) -| node[l,pos=0.75,rotate=-90,above] {2$\times$4K Video} ($(rendering.north east)+(-0.4,0)$);
  \draw[latex-] ($(head.north east)+(0,-0.4)$) -| node[l,pos=0.75,rotate=90,above] {Head Pose} ($(rendering.north west)+(+0.4,0)$);

  \node[anchor=south,fill=yellow!50,draw=black] at (head.south) {Robot with 6D Head};

  \node[anchor=south,fill=yellow!50,draw=black] at (vr.south) {Stereoscopic VR System};

  \node[below=0.7cm of rendering] (fast) {11\,ms};
  \draw[-latex] (fast) ++(0.6cm,0) arc (0:-330:0.6cm);

  \node[above=0.6cm of rendering] (slow) {150\,ms};
  \draw[-latex] (slow) ++(0.6cm,0) arc (0:-330:0.6cm);
 \end{tikzpicture}
 \caption{Our low-latency televisualization approach. The 2$\times$4K video stream from a stereo pair of wide-angle cameras is rendered using a sphere model to the operator's VR headset.
 On a fast timescale, operator movements are handled without much latency by re-rendering from the new pose.
 The operator's pose is also streamed to the robotic system, which moves the cameras accordingly in 6D, although on a slower timescale.
 }
 \label{fig:teaser}
\end{figure}

\section{Related Work}

\begin{figure*}
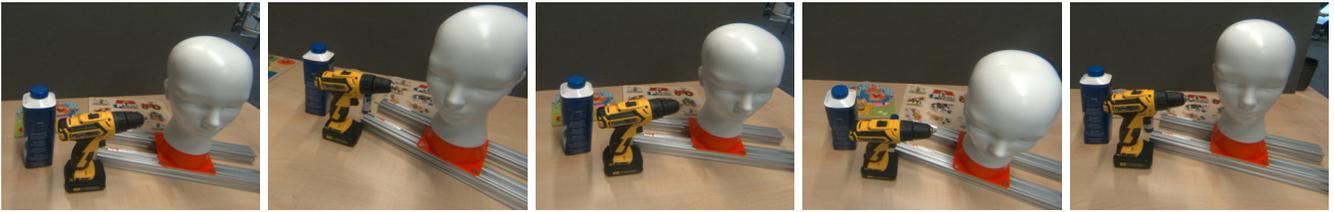

 \centering
 \includegraphics[height=2.75cm,clip,trim=500 200 0 600]{images/perspective/mpv-shot0003.jpg}
 \includegraphics[height=2.75cm,clip,trim=500 200 0 600]{images/perspective/mpv-shot0008.jpg}
 \includegraphics[height=2.75cm,clip,trim=500 200 0 600]{images/perspective/mpv-shot0011.jpg}
 \includegraphics[height=2.75cm,clip,trim=500 200 0 600]{images/perspective/mpv-shot0016.jpg}
 \includegraphics[height=2.75cm,clip,trim=500 200 0 600]{images/perspective/mpv-shot0017.jpg}
 \caption{Range of motion supported by our setup. We show the left eye view of the HMD, cropped
   to the central part for clarity. Note that the human operator can easily change
   perspective and observe parts of the scene that were previously hidden.}
 \label{fig:perspective}
\end{figure*}

Viewing a stereo camera video feed is one of the established methods for remote immersion,
providing live imagery with intuitive depth perception.

\citet{martins2015design} mount a stereo camera on a field robot developed for rescue applications.
While the camera is fixed to the robot, its track-driven body can bend to adjust the camera pitch angle.
The operator wears a head-mounted display, whose tracking pose is used to control the pitch angle and the
yaw angle of the entire robot. Roll motions are simulated by rotating the camera images in 2D.
The authors report that the high latency on yaw changes was confusing for operators.
We also note that this method will not accurately represent head orientations due to the decomposition
into Euler angles, which are then applied at different locations.
In contrast, our method supports full 6D head movement (see \cref{fig:perspective}) and reacts to pose changes with almost
zero latency.

\citet{zhu2010head} study whether head or gaze tracking is more suitable for camera orientation control.
However, they do not use a HMD but a 2D monitor for displaying camera images. Both head and gaze control
modes operate in a relative fashion, moving the camera in the indicated direction until the head/eye returns
to its neutral position. The authors conclude that the gaze-based control method feels more natural.
In contrast, our system uses the absolute head pose to directly control the camera pose, leading to very
natural and intuitive control of the camera.

This technique can also be seen in the work of \citet{agarwal2016imitating}, who control a robot's head pose
directly from a human head pose measured with a Microsoft Kinect 2. Their robot head is only capable of 2D
movement, pitch and yaw, so the human pose is mapped with a neural network to the lower-dimensional space.
The robot head is not equipped with cameras, however.

Very similarly, in one of the earliest works we could find, \citet{heuring1996visual} control a stereo camera pair
mounted on a pan/tilt unit with sufficiently low latency to follow human head motions precisely.

Conversely, \citet{lipton2017baxter} combine stereo capture with VR to enable teleoperation, but mount their
camera in a fixed location on the robot, providing only limited field of view.

When the robot or its head cannot be moved fast enough, latencies are introduced into a real-time teleoperation system. One particularly
effective approach for removing latencies is to display not the live camera feed, but a reconstructed 3D scene, in which
the operator can move freely without much latency.
We explored this technique in our own work, where we rendered
reconstructed 3D laser maps of the environment to the user \citep{rodehutskors2015intuitive} and displayed
textured meshes generated from a live RGB-D feed \citep{klamt2020remote}. In both cases, we used a VR headset which
allowed the user free movement inside the reconstructed data with low latency.
Later on, we improved on these techniques in \citet{stotko2019vr} by displaying a high-fidelity 3D mesh of the
environment generated from an automatically moving RGB-D camera mounted on the end-effector of a mobile manipulator.

Visualizing reconstructed scenes has the disadvantage that although the system reacts to observer pose changes with
almost no latency (requiring just a rendering step), changes in the scene itself typically only show up after longer
delays---if the scene representation supports dynamic content at all. This makes these approaches completely
unsuited for many telepresence situations, e.g. when interacting with a person or performing manipulation.
Furthermore, reconstruction methods typically rely on depth, either measured by depth cameras or inferred (SfM).
Many materials and surfaces encountered in everyday scenes result in unreliable depth and may violate
assumptions made by the reconstruction method (e.g. transparent objects, reflections), resulting in wrong transformations
or missing scene parts.

RGB-D-based visualization solutions as the one by \citet{whitney2020comparing},
or \citet{sun2020new},
who display live RGB-D data
as point clouds for teleoperation of a Baxter robot, can cope with dynamic scenes but still suffer from
measurement problems and the inherent sparsity---the operator can often look
through objects.

Displaying a live stereo camera feed as done in our method neatly sidesteps these issues, since the 3D reconstruction happens
inside the operator's visual cortex. We address the resulting
higher observer pose latency using a smart rendering method discussed in the next section.

\section{Method}

\subsection{Robotic Platform \& VR Equipment}

Our robot's head is mounted on a UFACTORY xArm 6, providing full 6D control of the head (see \cref{fig:teaser,fig:calibframes}).
The robotic arm is capable of moving a 5\,kg payload, which is more than enough to position a pair of cameras
and a small color display for telepresence (not used in this work). Furthermore, the arm is very slim,
which results in a large workspace while being unobtrusive. Finally, it is capable of fairly high speeds
(180\,°/s per joint, roughly 1\,m/s at the end-effector), thus being able to match dynamic human head movements.

We chose two Basler a2A3840-45ucBAS cameras for the stereo pair, which offer 4K video streaming at 45\,Hz.
The cameras are paired with C-Mount wide-angle lenses, which provide more than 180° field of view
in horizontal direction. We also experimented with Logitech BRIO webcams with wide-angle converters,
which offer auto-focus but can only provide 30\,Hz at 4K, resulting in visible stutters with moving objects.
The Basler cameras are configured with a fixed exposure time (8\,ms) to reduce motion blur to a minimum.

The robot head is mounted on a torso equipped with two Franka Emika Panda arms and anthropomorphic hands.

The operator wears an HTC Vive Pro Eye head-mounted display, which offers 1440$\times$1600 pixels per eye with an update
rate of 90\,Hz and 110° diagonal field of view. While other HMDs with higher resolution and/or FoV exist,
this headset offers eye tracking which will be beneficial for building telepresence solutions in future work.
As can be seen, the camera FoV is much larger than the HMD FoV, which ensures that the visible FoV stays inside
the camera FoV even if the robot head should be lagging behind the operator's movement.

The VR operator station is driven by a standard PC (Intel i9-9900K 3.6\,GHz, NVidia RTX 2080), connected
via Gigabit Ethernet to the robot computer (same specifications).

\subsection{Calibration}

\begin{figure}
 \centering
 \includegraphics[width=\linewidth]{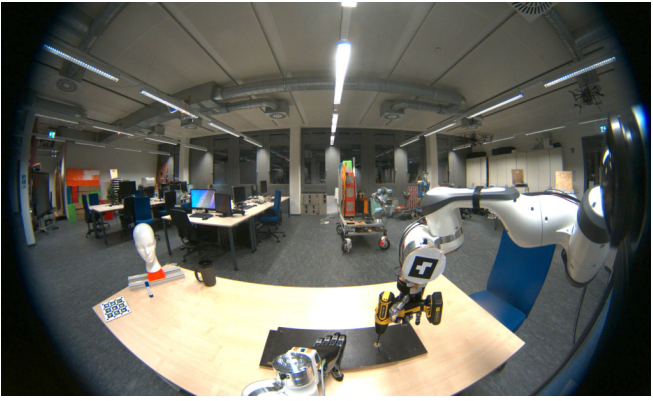}
 \caption{Raw image from one of the wide-angle cameras. The horizontal field of view is higher than 180°.}
 \label{fig:camera_image}
\end{figure}

Before the system can be used, multiple transforms and parameters need to be calibrated. We devised a principled
approach starting at camera intrinsic calibration over hand-eye calibration on the robot side, to VR calibration
on the operator side.

Since the cameras have very high FoV with significant fish-eye distortion (see \cref{fig:camera_image}),
we use the Double-Sphere camera model~\citep{usenko2018double} to describe the intrinsics. Its parameter sets $\mu$ and $\gamma$ for the left and right cameras, respectively, can be calibrated easily using the kalibr software package~\citep{rehder2016extending}, just requiring
an Aprilgrid calibration target. The kalibr software also estimates the extrinsic transformation between the two
cameras $T^L_R$.

\begin{figure}\centering
 \begin{tikzpicture}[
    clip,
    l/.style={fill=yellow!30, rounded corners,draw=black},
    o/.style={l,fill=blue!30},
  ]
  \node [opacity=0.3] (img) {\includegraphics[width=.75\linewidth]{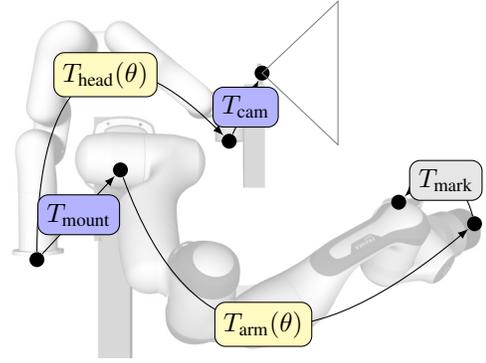}};

  \begin{scope}[x={($ (img.south east) - (img.south west) $ )},y={( $ (img.north west) - (img.south west)$ )}, shift={(img.south west)}]
   \node [circle,fill=black,inner sep=2pt] (headmount) at (0.085,0.3) {};
   \node [circle,fill=black,inner sep=2pt] (head) at (0.465,0.63) {};
   \node [circle,fill=black,inner sep=2pt] (cam) at (0.53,0.82) {};
   \node [circle,fill=black,inner sep=2pt] (armmount) at (0.25,0.55) {};
   \node [circle,fill=black,inner sep=2pt] (hand) at (0.95,0.4) {};
   \node [circle,fill=black,inner sep=2pt] (marker) at (0.8,0.46) {};

   \draw[-latex] (headmount) to[out=90,looseness=2.0] node[l,pos=0.6] {$T_\text{head}(\theta)$} (head);
   \draw[-latex] (head) -- node[o,midway] {$T_\text{cam}$} (cam);

   \draw[-latex] (headmount) -- node[o,midway] {$T_\text{mount}$} (armmount);

   \draw[-latex] (armmount) to[out=-70,in=-140,looseness=1.5] node[l,midway] {$T_\text{arm}(\theta)$} (hand);

   \draw[-latex] (hand) to[bend right=60,looseness=1.5] node[o,fill=gray!20,midway] {$T_\text{mark}$} (marker);

    \draw[draw=black!50] (cam.center) -- ++(0.15,0.2) -- ++(0,-0.4) -- cycle;
  \end{scope}
 \end{tikzpicture}
 \caption{Frames involved in the calibration procedure. Transformations colored yellow are computed using
   forward kinematics from measured joint angles while blue transformations are optimized.
   The auxiliary transform $T_\text{marker}$ is optimized, but not used in the calibrated system.
   Note: The camera FoV is shown for illustration purposes and is not to scale.}
 \label{fig:calibframes}
\end{figure}

As shown in \cref{fig:calibframes}, there are further extrinsic transformations to be calibrated, especially if one wants to accurately perform manipulation
with a robotic arm. We follow a classic
hand-eye calibration approach with an ArUco marker on the robot's wrist, with the goal of estimating $T_{\text{cam}}$, but also the transformation $T_{\text{mount}}$ between the mounting of the head arm and the robot arm, and $T_{\text{mark}}$, the pose of the marker on the wrist.

We can compute the 2D pixel positions $p^L$ and $p^R$ in the left and right camera image frame as follows:
\begin{alignat}{2}
 p^L &= f_\mu(T_{\text{cam}}^{-1} \cdot T_\text{head}^{-1}(\theta) \cdot T_{\text{mount}} \cdot T_\text{arm}(\theta) \cdot T_\text{mark}), \\
 p^R &= f_\gamma(T^L_R \cdot T_{\text{cam}}^{-1} \cdot T_\text{head}^{-1}(\theta) \cdot T_{\text{mount}} \cdot T_\text{arm}(\theta) \cdot T_\text{mark}),
\end{alignat}
where $T_{\text{arm}}(\theta)$ and $T_{\text{head}}(\theta)$ are the poses of the arm and head flanges relative to their bases for the current joint configuration $\theta$, and $f$ is the double-sphere projection function with parameters $\mu$ and $\gamma$ found above.

We collect samples with known 2D pixel coordinates $\hat{p}^L$ and $\hat{p}^R$ extracted using the ArUco marker detector of the OpenCV library.
During sample collection, the head continuously moves in a predefined sinusoidal pattern while the robot's arm
is moved manually using teach mode. Sample collection takes about 5\,min for an engineer.

Finally, we minimize the squared error function
\begin{alignat}{2}
 E = \sum_{i=1}^N ||p^L - \hat{p}^L||_2^2 + ||p^R - \hat{p}^R||_2^2
\end{alignat}
over all $N$ samples using the Ceres solver, yielding optimal transforms $T_{\text{cam}}$ and $T_{\text{mount}}$.

\subsection{Head Control}

The remaining question is how to transfer the head pose, measured in the VR space, to the robot side. To initialize
the mapping, we keep the robot head in predefined nominal pose $T^\text{robot}_\text{nom}$, looking straight ahead. We ask the
human operator to do the same and record the resulting head pose in VR space $T^\text{VR}_\text{nom}$, taking care to remove any remaining undesirable pitch and roll component.

During operation, the head pose $T^\text{robot}_\text{head}$ is computed from the tracking pose $T^\text{VR}_\text{head}$ in straightforward fashion:
\begin{alignat}{2}
 T^\text{robot}_\text{head} = T^\text{robot}_\text{nom} \cdot (T^\text{VR}_\text{nom})^{-1} \cdot T^\text{VR}_\text{head},
\end{alignat}
directly mapping the operator's head movement to the robot's head.

The head pose is smoothed by a recursive low-pass filter with a cut-off frequency of 100\,Hz to remove very high frequency
components. It is also monitored for larger jumps, in which case we switch to a slower initialization mode that moves to
the target pose with limited velocity.

\subsection{Spherical Rendering}

If operator and robot head poses matched exactly, the camera images could be rendered immediately to the HMD by projecting
each pixel in HMD image space to the camera images and looking up the corresponding color value.
However, when the poses are not identical, as is usually the case, we have a 6D transform $T^V_C$ between the virtual
VR camera $V$ (representing the operator's eye) and actual camera position $C$.
Traversing this transform requires 3D positions, but we do not have depth information.

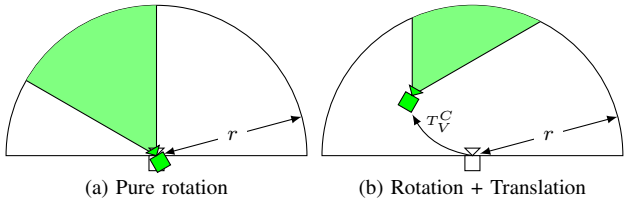
\begin{figure}
 \centering
 \begin{tikzpicture}[font=\footnotesize]
  \begin{scope}
   \begin{scope}
    \draw (2,0) arc (0:180:2) -- cycle;
    \clip (2,0) arc (0:180:2) -- cycle;

    \begin{scope}[rotate=30]
      \draw[fill=green!50]
        (0,0) -- (60:3) arc [radius=3,start angle=60, delta angle=60] -- cycle;
    \end{scope}

    \draw[latex-latex] (15:0.1) -- node[midway,fill=white] {$r$} (15:2);
   \end{scope}
   \draw[fill=white]
     (-0.1,-0.2) rectangle (0.1,0.0)
     (0.0,0.0) -- ++(-0.1,0.1) -- ++(0.2,0.0) -- cycle;

   \begin{scope}[rotate=30]
   \draw[fill=green]
     (-0.1,-0.2) rectangle (0.1,0.0)
     (0.0,0.0) -- ++(-0.1,0.1) -- ++(0.2,0.0) -- cycle;
   \end{scope}
   \node[anchor=north] at (0,-0.2) {(a) Pure rotation};
  \end{scope}
  \begin{scope}[shift={(4.2,0)}]
   \begin{scope}
    \draw (2,0) arc (0:180:2) -- cycle;
    \clip (2,0) arc (0:180:2) -- cycle;

    \begin{scope}[shift={(-0.8,0.8)}, rotate=-30]
       \draw[fill=green!50]
       (0,0) -- (60:4) arc [radius=4,start angle=60, delta angle=60] -- cycle;
    \end{scope}
    \draw[latex-latex] (15:0.1) -- node[midway,fill=white] {$r$} (15:2);
   \end{scope}
   \draw[fill=white]
       (-0.1,-0.2) rectangle (0.1,0.0)
       (0.0,0.0) -- ++(-0.1,0.1) -- ++(0.2,0.0) -- cycle;

   \begin{scope}[shift={(-0.8,0.8)}, rotate=-30]
   \draw[fill=green]
       (-0.1,-0.2) rectangle (0.1,0.0)
       (0.0,0.0) -- ++(-0.1,0.1) -- ++(0.2,0.0) -- cycle;
   \end{scope}

   \draw[-latex] (0,0) to [in=-60,out=170] node [pos=0.9,right,font=\tiny] {$T^C_V$} (-0.8,0.55);

   \node[anchor=north] at (0,-0.2) {(b) Rotation + Translation};
  \end{scope}
 \end{tikzpicture}
 \vspace{-2ex}
 \caption{%
   Spherical rendering example in 2D. We show only one camera $C$ of the stereo pair,
   the other is processed analogously.
   The robot camera is shown in white with its very wide FoV. The corresponding VR camera $V$,
   which renders the view shown in the HMD, is shown in green.
   The camera image is projected onto the sphere with radius $r$, and then back into the VR camera.
   Pure rotations (a) result in no distortion, while translations (b) will
   distort the image if the objects are not exactly at the assumed depth.
 }
 \label{fig:spherical}
\end{figure}

Hence, we must make assumptions which will not impact the result quality much, since the transform $T^V_C$ will
be quite small.
The simplest assumption is one of constant depth for the entire scene. Since we are dealing with wide-angle cameras, it
is natural to assume constant \textit{distance}, leading to spherical geometry around the camera.
With this assumption, we can raycast from the VR camera $V$, find the intersection with the sphere, and find the corresponding pixel in the camera image (see \cref{fig:spherical}).

In practice, the rendering part is implemented in OpenGL using the Magnum 3D Engine\footnote{\url{https://magnum.graphics}}.
We use the standard rasterization pipeline by rendering a sphere of radius $r=1\,\text{m}$ at the camera location.
A fragment shader then computes the pixel location (in $C$) of each visible point of the sphere and performs a texture
lookup with bilinear interpolation to retrieve the color value.

\begin{figure}
 \begin{tikzpicture}[font=\footnotesize]
  \begin{axis}[height=4cm,width=\linewidth,xmin=0,xmax=3,ymin=-10,xlabel={Real object distance $d$ [m]},ylabel={Angular Error $\gamma$ [$^\circ$]}]
   \draw[very thin] (axis cs:0,0) -- (axis cs:5,0);

   \addplot+[no marks,thick] [domain=0.2:3,samples=50] { (atan(x/0.1) - atan(1/0.1)) };
   \addplot+[no marks,dashed] [domain=0:3,samples=2] { (90 - atan(1/0.1)) };

  \end{axis}
 \end{tikzpicture}
 \vspace{-2ex}
 \caption{Angular error introduced by translation of the operator head by $\Delta x = 10$\,cm depending on the distance to the object.
  Since the spherical reprojection assumes a distance of 1\,m, we see no error at that distance.
  The red dashed line shows the error upper bound for large distances $d$.}
 \label{fig:angular_error}
\end{figure}
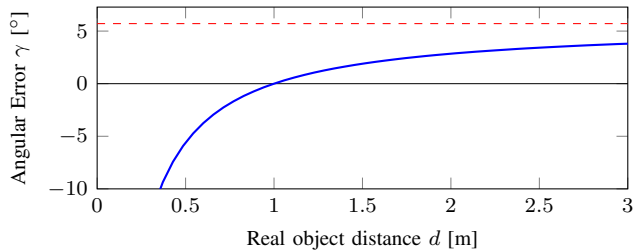

Using this design, head movements are immediately reacted to. Pure rotations of the head will have no perceptible
latency, since the wide-angle cameras will already have captured the area the operator will look at.
Translational movements work as if everything is at a constant distance, yielding distortions wherever this assumption does not hold (see \cref{fig:spherical}). The amount of distortion can be quantified by the angular projection error $\gamma$ and depends on the distance $d$ to the object and the head translation $\Delta x$ (here perpendicular to the view ray):

\begin{equation}
 \gamma = \arctan \bigg (\frac{d}{\Delta x} \bigg ) - \arctan \bigg(\frac{r}{\Delta x}\bigg).
\end{equation}

Note that $\gamma$ quickly saturates for large $d$ against the value $\frac{\pi}{2} - \arctan (\frac{r}{\Delta x})$ (see \cref{fig:angular_error}). For $d < r$, the distortion can become significant.
In conclusion, it is important to choose $r$ close to a typical minimum object distance.
Of course, as soon as the robot head catches up and moves into the correct pose, any distortion is corrected.

\subsection{Latency Improvements}

We found that we needed to take special care at nearly all parts of the pipeline to ensure low-latency operation
with the required time precision.

Starting at the cameras, we wrote a custom driver which uses the built-in timestamping feature of the cameras
to obtain high-precision timing information in order to correlate images with head-arm transforms.
Furthermore, it is necessary to compress the image stream for transmission over Ethernet, since the raw stream
bandwidth exceeds 350\,MB/s. For easy interoperability  with standard ROS packages, we chose an MJPEG compression,
which is done efficiently on the GPU~\citep{holub2012ultragrid}.
The resulting stream has a bandwidth of 30\,MB/s, which is acceptable in our setup. If even less bandwidth
is required, hardware-accelerated H.264 encoding could be a solution.

Furthermore, we developed a custom driver for the xArm, which provides joint position measurements at a high
rate. This required modifications to the arm firmware. Our modified version can control and report
joint positions at 100\,Hz. Its measurements are interpolated and
synchronized with the timestamped camera images using the ROS \texttt{tf} stack.

Finally, the VR does not use the standard \texttt{image\_transport} ROS package, which
cannot decompress the two MJPEG streams with 4K resolution at 45\,Hz.
Instead, we use the GPUJPEG decoder~\citep{holub2012ultragrid}.

All in all, these examples highlight that many off-the-shelf products are not designed for the required
low-latency operation and manual adaptation requires considerable effort.

\section{Experiments}
\label{sec:eval}

\subsection{Quantitative Experiments}

In a first experiment, we measured the total image latency from acquisition start inside the camera
to display in the HMD. For this, we used the timestamping feature of the camera, with the camera time synchronized
to the robot computer time. Both computers are synchronized using the \textit{chrony} NTP system, so that timestamps
are comparable. Finally, the VR renderer compares the system time against the camera timestamp at rendering each frame.
On average, the images are shown with 40\,ms latency. A considerable fraction of this time is consumed by
the camera exposure time
(8\,ms) and USB transfer time (around 10\,ms).

\pgfplotstableread{data/latency.log}\latencytable
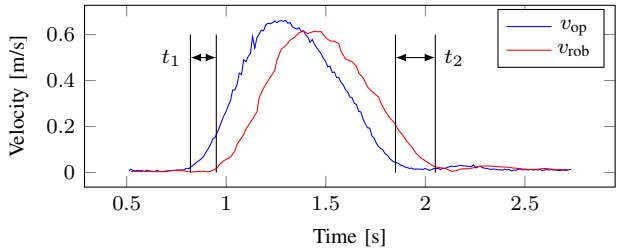
\begin{figure}
  \centering
  \begin{tikzpicture}[font=\footnotesize]
   \begin{axis}[height=4cm,width=\linewidth,
     unbounded coords=discard,
     filter discard warning=false, xlabel={Time [s]}, ylabel={Velocity [m/s]},
     legend entries={$v_{\text{op}}$,$v_{\text{rob}}$}]
     \addplot+ [mark=none] table [x expr={\thisrow{time}-40},y=ctrans] {\latencytable};
     \addplot+ [mark=none] table [x expr={\thisrow{time}-40},y expr=\thisrow{mtrans}/2,legend] {\latencytable};

     \draw (1.85,0) -- (1.85,0.6);
     \draw (2.05,0) -- (2.05,0.6);
     \draw[latex-latex] (1.85,0.5) -- (2.05,0.5) node [anchor=west] {$t_2$};

     \draw (0.82,0) -- (0.82,0.6);
     \draw (0.95,0) -- (0.95,0.6);
     \draw[latex-latex] (0.82,0.5) node [anchor=east] {$t_1$} -- (0.95,0.5);
   \end{axis}
  \end{tikzpicture}
  
  \vspace{-2ex}
  
  \caption{Head movement latency experiment. Shown are the Cartesian velocities of the operator's head $v_{\text{op}}$
  and the robot's head $v_{\text{rob}}$ during a fast (0.5\,m/s) head movement. Vertical lines have been added to ease analyzing latencies (start latency $t_1 \approx 130$\,ms, finish latency $t_2 \approx 100$\,ms).}
  \label{fig:head_latency}
\end{figure}

Since the VR renderer achieves the maximum HMD update rate of 90\,Hz without difficulty, rotational head movements
have the same latency as in local VR applications, which is not noticeable---according to publicly available information,
the Lighthouse VR tracking system operates with 250\,Hz update rate\footnote{\url{https://en.wikipedia.org/wiki/HTC_Vive}}.

\pgfplotstableread{data/camdiffsub.log}\camdifftable
\begin{figure}
 \centering
  \begin{tikzpicture}[font=\footnotesize]
   \pgfplotsset{set layers}

   \begin{axis}[
     scale only axis,
     height=2cm,width=.7\linewidth,
     xlabel={Time [s]}, ylabel={Deviation [cm]},
     legend entries={$\Delta s$,$v_{\text{op}}$},
     legend pos=north west ,
     axis y line*=left,
     xmin=1, xmax=16,
     ]
     \addplot+ [blue,mark=none] table [x expr={\thisrow{time}-15},y expr=100*\thisrow{camdiff}] {\camdifftable};
     \addplot+ [red,mark=none] coordinates {(0,0)};
   \end{axis}
   \begin{axis}[
     scale only axis,
     height=2cm,width=.7\linewidth,
     xlabel={Time [s]}, ylabel={Velocity [cm/s]},
     axis x line=none,
     axis y line*=right,
     xmin=1, xmax=16,
     ]
     \addplot+ [red,mark=none] table [x expr={\thisrow{time}-15},y expr=50*\thisrow{ctrans}] {\camdifftable};
   \end{axis}
  \end{tikzpicture}
  \vspace{-.2cm}
  \caption{Head tracking performance. We show the distance $\Delta s$ of the VR eye camera $V$ from the robot camera $C$ for the
  left eye stream.
  The operator was performing highly dynamic head movements.
  As expected, the position deviation is highly correlated with operator head velocity $v_{\text{op}}$.
  }
  \label{fig:head_deviation}
  \vspace{-2ex}
\end{figure}
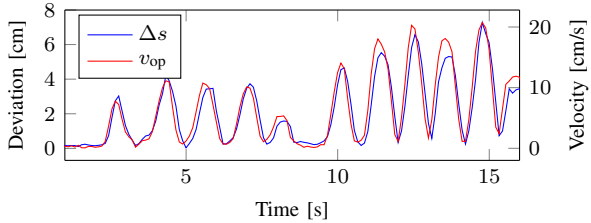

To analyze the total latency of  head control, we plotted Cartesian velocities of the raw target head pose (as
recorded by the VR tracking system) and the measured head pose during a fast head motion (see \cref{fig:head_latency}).
The latency varies along the trajectory, but the head comes to a stop roughly 200\,ms after the operator's head stopped moving. We note that this is a maximum bound of the latency; slower head motions result in lower latencies.

Furthermore, we analyzed the absolute deviation between the HMD eye pose and the camera pose, i.e. the error that
is introduced during the spherical rendering step (see \cref{fig:head_deviation}). We note that even during
fast motions, only minor distortions were noticeable to the operator.

\subsection{User Study}

We conducted a small user study in order to measure the immersiveness and efficacy of our pipeline. Due to
the ongoing COVID-19 pandemic, we were limited to immediate colleagues as subjects, which severely constrained
the scope of our study to seven participants.

\begin{figure}
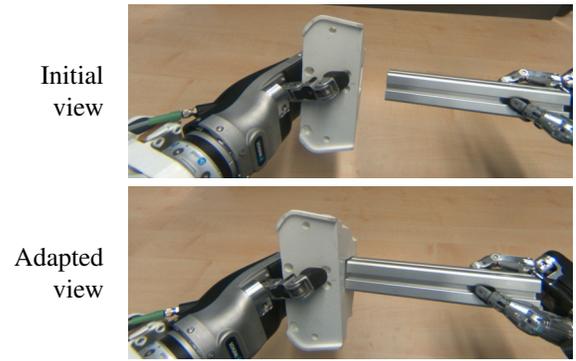

 \centering
 \begin{tikzpicture}
  \node[inner sep=0cm] (img1) {\includegraphics[height=2.3cm,clip,trim=100 300 0 800]{images/perspective/mpv-shot0018.jpg}};
  \node[left=0.2cm of img1,align=right] (label1) {Initial\\view};
  
  \node[below=0.1cm of img1,inner sep=0cm] (img2) {\includegraphics[height=2.3cm,clip,trim=100 300 0 800]{images/perspective/mpv-shot0019.jpg}};
  \node[left=0.2cm of img2,align=right] (label1) {Adapted\\view};
 \end{tikzpicture}
 \caption{%
   Operator view (cropped) during the user study task. The peg has square cross section of 4$\times$4\,cm.
   The hole in the white plastic object is 0.3\,mm larger.
   In the bottom situation, the user repositioned their head to better view the interfacing parts.
 }
 \label{fig:task}
\end{figure}

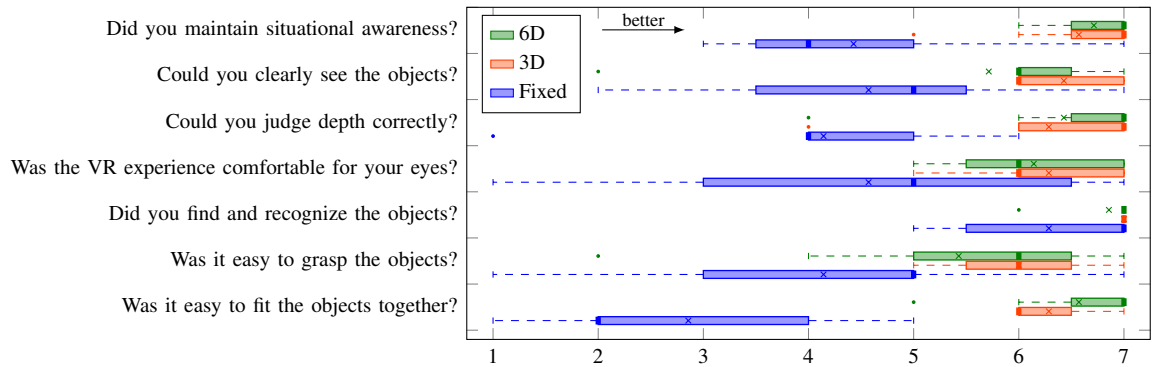
\begin{figure*}
 \centering
\begin{tikzpicture}[font = \footnotesize, every mark/.append style={mark size=0.5pt}]
 \begin{axis}[
     name=plot,
     boxplot/draw direction=x,
     width=0.6\textwidth,
     height=6cm,
     boxplot={
         draw position={1.5 - 0.325/3 + 1.0*floor((\plotnumofactualtype + 0.001)/3) + 0.2*mod((\plotnumofactualtype + 0.001),3)},
         box extend=0.17,
         average=auto,
         every average/.style={/tikz/mark=x, mark size=1.5, mark options=black},
         every box/.style={draw, line width=0.5pt, fill=.!40!white},
         every median/.style={line width=2.0pt},
         every whisker/.style={dashed},
     },
     ymin=1,
     ymax=8.2,
     y dir=reverse,
     ytick={1,2,...,13},
     y tick label as interval,
     yticklabels={
Did you maintain situational awareness?,
Could you clearly see the objects?,
Could you judge depth correctly?,
Was the VR experience comfortable for your eyes?,
Did you find and recognize the objects?,
Was it easy to grasp the objects?,
Was it easy to fit the objects together?
     },
     y tick label style={
         align=center
     },
     xmin=0.75,
     xmax=7.25,
     xtick={1, 2 ,..., 7},
     xticklabels = {1, 2, ..., 7},
     cycle list={{green!50!black,orange!50!red,blue}},
     y dir=reverse,
     legend image code/.code={
         \draw [#1, fill=.!40!white] (0cm,-1.5pt) rectangle (0.3cm,1.5pt);
     },
     legend style={
         anchor=north west,
         at={($(0.0,1.0)+(0.2cm,-0.1cm)$)},
     },
     legend cell align={left},
 ]

 \addplot
 table[row sep=\\,y index=0] {
 data\\
 7\\7\\7\\6\\7\\7\\6\\
 };

 \addplot
 table[row sep=\\,y index=0] {
 data\\
 7\\7\\7\\7\\7\\5\\6\\
 };

 \addplot
 table[row sep=\\,y index=0] {
 data\\
 3\\5\\5\\7\\4\\3\\4\\
 };

 \addplot
 table[row sep=\\,y index=0] {
 data\\
 6\\6\\2\\6\\7\\7\\6\\
 };

 \addplot
 table[row sep=\\,y index=0] {
 data\\
 7\\6\\6\\6\\7\\7\\6\\
 };

 \addplot
 table[row sep=\\,y index=0] {
 data\\
 2\\4\\5\\6\\7\\5\\3\\
 };

 \addplot
 table[row sep=\\,y index=0] {
 data\\
 7\\4\\7\\6\\7\\7\\7\\
 };

 \addplot
 table[row sep=\\,y index=0] {
 data\\
 7\\4\\7\\6\\7\\7\\6\\
 };

 \addplot
 table[row sep=\\,y index=0] {
 data\\
 1\\4\\5\\6\\4\\4\\5\\
 };

 \addplot
 table[row sep=\\,y index=0] {
 data\\
 6\\6\\5\\5\\7\\7\\7\\
 };

 \addplot
 table[row sep=\\,y index=0] {
 data\\
 6\\6\\5\\6\\7\\7\\7\\
 };

 \addplot
 table[row sep=\\,y index=0] {
 data\\
 1\\2\\5\\6\\7\\4\\7\\
 };

 \addplot
 table[row sep=\\,y index=0] {
 data\\
 7\\6\\7\\7\\7\\7\\7\\
 };

 \addplot
 table[row sep=\\,y index=0] {
 data\\
 7\\7\\7\\7\\7\\7\\7\\
 };

 \addplot
 table[row sep=\\,y index=0] {
 data\\
 5\\7\\7\\7\\6\\7\\5\\
 };

 \addplot
 table[row sep=\\,y index=0] {
 data\\
 6\\2\\4\\7\\7\\6\\6\\
 };

 \addplot
 table[row sep=\\,y index=0] {
 data\\
 6\\7\\5\\6\\7\\5\\6\\
 };

 \addplot
 table[row sep=\\,y index=0] {
 data\\
 1\\7\\5\\5\\2\\4\\5\\
 };

 \addplot
 table[row sep=\\,y index=0] {
 data\\
 7\\5\\6\\7\\7\\7\\7\\
 };

 \addplot
 table[row sep=\\,y index=0] {
 data\\
 6\\6\\7\\6\\7\\6\\6\\
 };

 \addplot
 table[row sep=\\,y index=0] {
 data\\
 1\\2\\5\\4\\2\\4\\2\\
 };

 \legend{6D,3D,Fixed}

 \end{axis}

 \draw[-latex] ($(plot.north west)+(1.8cm,-0.3cm)$) -- ++(1.1cm,0cm)
 node[midway,above,font=\scriptsize,inner sep=1pt] {better};

 \end{tikzpicture}
  \vspace{-0.5cm}
 \caption{
   Statistical results of our user questionnaire. We show the median, lower and upper quartile (includes interquartile range), lower and upper fence, outliers (marked with •) as well as the average value (marked with $\times$), for each aspect as recorded in our questionnaire.
  }
  \label{fig:userstudy}
  \vspace{-2ex}
\end{figure*}

Participants were asked to teleoperate the robot, performing a rather difficult peg-in-hole task involving grasping two objects, and inserting one into the other with tight tolerances (see \cref{fig:task}).
The task places high demands on the visual system
since a) even small misalignments would prevent success and b) the peg object was quite hard to hold and slippage
could only be detected visually.
For controlling the two robot arms and hands, the participants used a custom force-feedback exoskeleton system which is not the focus of this work.
The participants performed the task three times, with different visual modes:

\paragraph{Full 6D}
Entire pipeline with full 6D movement.

\paragraph{3D orientation}
The operator only controls the orientation of the head (similar to \citep{heuring1996visual,martins2015design}).
The spherical rendering system is used, but the camera spheres are translated to the HMD eye position.
Notably, this variant can still react to orientation changes with low latency by re-rendering.

\paragraph{Fixed perspective}
The robot head orientation is fixed to a 45° downward pitch, focusing on the table in front of the robot.
The camera images are rendered directly to the HMD, i.e. the HMD pose has no effect. This is similar
to devices sold as First-Person-View (FPV) glasses, which are commonly used in drone racing.

The order of the three trials was randomized to mitigate effects from learning during the trials.
Participants could attempt the task as often as they liked (with an assisting human resetting the
objects on the table if moved out of reach). When a maximum time of 5\,min was reached without success,
the trial was marked as failed.

\begin{table}
 \centering
 \caption{User study success rates and timings}\label{tab:success}
 \vspace{-2ex}
 \begin{tabular}{lrrr}
  \toprule
  Visual mode & Success & \multicolumn{2}{c}{Completion time [s]} \\
  \cmidrule (lr) {3-4}
              &         & \hspace{2em}Mean & StdDev \\
  \midrule
  \textbf{a) Full 6D} & \textbf{7/7} & \textbf{71.0} & \textbf{50.6} \\
  b) 3D orientation   & 7/7 & 111.7 & 87.5 \\
  c) Fixed perspective & 6/7 & 158.3 & 39.6 \\
  \bottomrule
 \end{tabular}
 \vspace{-2ex}
\end{table}

Although the task itself was not the primary goal of the user study, effects
of the visual mode on success rates and average completion time can be observed (\cref{tab:success}).
The full 6D mode never led to a failure and yielded the lowest average completion time, although the standard deviation was quite high---mostly
due to external effects such as the robot arms requiring a reset after too much force was applied, but the skill level also varied highly from participant to participant.

After the three trials, we asked the participants questions for each of the visualization modes, with answer
possibilities from the 1-7 Likert scale.
From the results reported in \cref{fig:userstudy},
it is immediately visible that the fixed perspective variant leads to significantly lower
user acceptance. The users reported that this was mainly due to inability to pan the camera and instead
having to move the workpieces to a better observable pose.
Furthermore, the head translation seems to have no impact on grasping the objects, but impacts
insertion---where it is beneficial to be able to look slightly from the side to see the involved
edges (compare \cref{fig:task}).

\section{Discussion \& Conclusion}

We have demonstrated a method for 3D remote visualization that does not rely on geometry reconstruction and is thus suited
for highly dynamic scenes, but is still able to deliver low-latency response to 6D head movements.
The user study shows that allowing full 6D head movement leads to faster task completion and better user acceptance
when compared with fixed-mount and pan-tilt systems, which are still common in teleoperation.

There are also some points of the system that can be improved.
The work space of the arm used to move the head is---in its present mounting position---limited and cannot match the full motion range of sitting humans moving their torso and neck. For this, it would be interesting to
investigate a more biologically inspired joint configuration.

While our method has been evaluated only in a seated configuration, it should be applicable with a completely mobile
robot and operator as well. This can be investigated in future research.

Our work also does not focus on the transmission of data to remote sites. It is clear that for transmission over the
Internet, bandwidth requirements would have to be lowered, e.g. through the use of more efficient video compression codecs.
The system should, however, be able to cope with typical latencies found
on Internet connections (up to 200\,ms for connections halfway around the world).

\printbibliography

\end{document}